**Title: Offline reinforcement learning with uncertainty for treatment strategies in sepsis**

**Authors:** Ran Liu, BS[1,2], Joseph L. Greenstein, PhD[1,2], James C. Fackler, MD[3], Jules Bergmann, MD[3], Melania M. Bembea, MD, MPH, PhD[3,4], Raimond L. Winslow, PhD[1,2*]

**Affiliations:**

[1]Institute for Computational Medicine, The Johns Hopkins University

[2]Department of Biomedical Engineering, The Johns Hopkins University School of Medicine & Whiting School of Engineering

[3]Department of Anesthesiology and Critical Care Medicine, and [4]Department of Pediatrics, The Johns Hopkins University School of Medicine

**Abstract:**

Establishing a guideline-based treatment for sepsis and septic shock remains challenging because sepsis is a disparate range of life-threatening organ dysfunctions whose pathophysiology is not fully understood. Early intervention in sepsis is crucial for patient outcome, yet these interventions have adverse effects and are frequently overadministered. Greater personalization is necessary, as no single action is suitable for all patients. We present a novel application of reinforcement learning in which we identify optimal recommendations for sepsis treatment from data, estimate their confidence level, and identify treatment options infrequently observed in training data. Rather than a single recommendation, our method presents several viable treatment options. We examine learned policies and discover that reinforcement learning is biased against aggressive intervention due to the confounding relationship between mortality and level of treatment received. We mitigate this bias using subspace learning, and develop methodology that can yield more accurate learning policies across healthcare applications.

**Introduction:**

Sepsis and septic shock incur the greatest in-hospital mortality[1] and cost of care among all conditions in the United States[2]. Consequently, they have been the subject of several decades of efforts to improve patient outcomes[3], which have led to a nearly 2-fold reduction in both mortality and cost in severe cases[4]. Sepsis is not a single disease, but a disparate range of life-threatening organ dysfunctions arising due to infection[5] whose pathophysiology is still not fully understood. The heterogeneity of sepsis makes treatment difficult, and many current guidelines still lack strong empirical support[6]. There is a strong association between treatment delays in septic shock and mortality[7-9], and therefore current clinical recommendations are focused primarily on immediate intervention[10]. To help addres this, a number of computational methods for early prediction of sepsis and septic shock with the aim of reducing treatment delays have been proposed[11-15].

Treatments for sepsis and septic shock can include fluid and vasopressor administration to increase blood pressure and enhance organ perfusion, and administration of antibiotics to fight infection. However, none of these interventions are without risk. Excessive fluid administration leads to overload in as many as two-thirds of sepsis patients[16]. Fluid overload increases risk of hypertension, pulmonary edema, and respiratory failure[16,17]. High dosages of vasopressors have been associated with increased mortality[18] and adverse cardiac events[19,20]. Antibiotics are frequently administered to patients without evidence of bacterial infection[21], resulting in antibiotic resistance and Clostridium difficile infection[20]. A greater degree of personalization of sepsis treatment is necessary, as no single policy is appropriate for all patients. Stratification and subtyping of sepsis patients toward this end is of interest[22]. In pursuit of this goal, we recently demonstrated a new method for and potential clinical benefits of clustering sepsis patients based on the evolution of their state and risk trajectories over time[23].

Reinforcement learning presents an avenue for learning treatment strategies from observations of physiological state, clinician actions, and consequent patient outcomes, by modeling patient state evolution and treatment responsiveness as a Markov decision process (MDP)[24]. Komorowski et al.[25] applied Q-learning to electronic health record (EHR) data to learn strategies for administering fluids and vasopressors to sepsis patients; the same group has applied additional deep reinforcement learning methods to this problem as well[26,27]. However, there has been relatively little follow-up in this area from other groups[28]. The fundamental challenge is that while many reinforcement learning methods can be trained on and benefit from historical data, further fine-tuning through online interactions (i.e., new data collected using the learned policies) is necessary to learn effective policies[29,30]. Online learning in healthcare may require ceding some decision-making for treatments of real patients from clinicians to artificial intelligence (AI), and therefore cannot be undertaken without careful consideration of additional associated patient risk. Some recent algorithmic advances in deep reinforcement learning have been aimed towards mitigating the technical challenges of learning effective policies through purely offline learning[31-33]. Rather than a purely algorithmic approach, however, we also examine learned policies from the clinical perspective, and compare AI recommendations to clinician actions, as well as established clinical best practices.

We present a novel application of reinforcement learning in which we compute confidence bounds on AI evaluations of sepsis interventions, identifying the confidence level of AI treatment recommendations. We also identify actions which are infrequently or never observed in the training data. This provides clinicians with more information and greater discretion than a black-box recommendation. In addition to decision algorithms based on terminal rewards determined by sepsis patient mortality, we explore intermediate reward formulations which are more similar to the short-term resuscitation objectives pursued by clinicians.

We also discover a clinically important problem in state-space formulation for reinforcement learning agents in healthcare applications: if heterogenous data from patients who respond differently to treatment are assigned to the same point in a discrete state-space (which is likely to occur with uniform binning approaches), the resulting AI will be biased towards recommending less intervention than is appropriate. This is because on average, healthier patients have better outcomes, despite receiving interventions at lower rates[23]. To mitigate this problem, we define a discrete state space by clustering on a learned subspace of physiological data that maximizes correlation with outcomes and interventions. This approach ensures a clinically meaningful partitioning of the state space in which states are more internally homogenous in terms of clinical labels, as well as response to intervention.

**Results:**

We used data from the MIMIC-III intensive care database (version 1.4)[34] to train a reinforcement learning agent for the administration of fluids and vasopressors in sepsis patients. Sepsis was determined according to the Third International Consensus Definitions for Sepsis and Septic Shock (Sepsis-3)[5], yielding 21,189 ICU stays from sepsis patients, and a total of 335,703 observed transitions (Table S1, S4-S6); each transition consists of a starting state, an action taken by the clinician, and an ending state.

Our fitted reinforcement learning agent produces time-evolving recommendations for the administration of fluids and vasopressors in sepsis patients (Figure 1). Each 4-hour time window is taken as a discrete time step. Fluid and vasopressor administrations are modeled as a discrete time, discrete state space MDP, where transition probabilities between states are conditioned upon actions taken at each discrete time step, and each transition has an associated reward value. An episode is the sequence of transitions observed over the course of a hospital admission. A policy is a distribution of actions for each state in the state space. The learned AI policy is deterministic (i.e., there is a single recommended action for each state), and chooses the action which maximizes the expected cumulative reward over all transitions in a given episode (i.e., the expected reward incurred from the current chosen action, plus the accumulated expected reward value from all subsequent actions).

Total fluid volume administered and the maximum dosage of vasopressors over each 4-hour window, as determined using vasoactive-inotropic score (VIS)[35], are binned into quintiles, based on the distribution of actions with non-zero dosages taken by clinicians in the dataset (Table 1). A zero dosage of vasopressors is treated as a separate bin, as the decision to initiate vasopressors is a clinically significant cutoff point. This yields a discrete action space of 30 actions, for each possible paired combination of vasopressors and fluids.

The state space was defined by using k-means clustering on physiological features, along with history variables containing clinician actions over the past 12 hours, extracted from EHR data (Table S2). Canonical correlation analysis (CCA)[36] was used for dimensionality reduction, and to learn a linear weighting of features that maximized correlation with interventions administered, clinical states, and patient outcomes prior to clustering. The number of states for k-means clustering was chosen using the elbow heuristic applied to the Akaike Information Criterion (AIC; Figure S1), yielding 200 clusters.

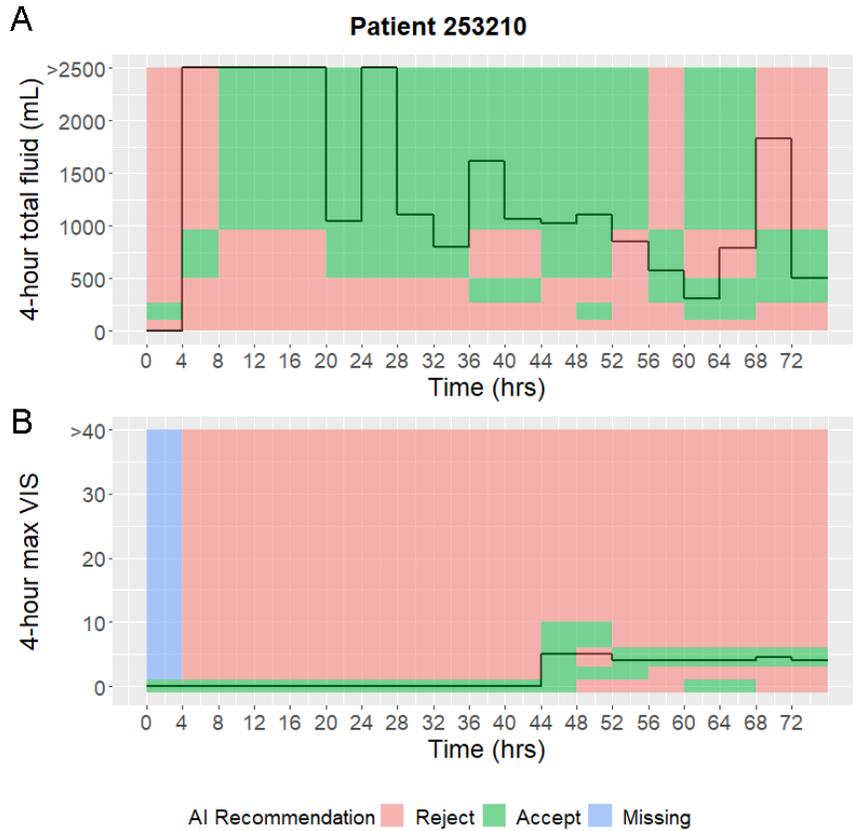

**Figure 1:** Clinician actions over the first 72 hours of an ICU admission, indicated by the black solid line, for total fluid administration (A) and maximum vasopressor dosage (B) over each 4-hour window. Ranges of actions rejected by the AI policy are indicated in red. Actions accepted by the AI policy are indicated in green. Actions unobserved in the dataset are indicated in blue. As multiple actions are represented for each bin of vasopressors or fluids alone (as for any fixed level of fluids, there are actions corresponding to multiple levels of vasopressors and vice versa), the action with the lowest p-value is given.

The Q-value of an action, given a particular starting state, is the cumulative expected reward incurred by that action. We estimated confidence bounds for the value of each action in each state using bootstrap; Figure 2 shows Q-values for state 48, which is occupied at hour 4 in Figure 1. In this case, reward takes on nonzero values only for transitions into the terminal states corresponding to mortality (reward = -1) or discharge (reward = +1), and thus a higher value corresponds to a lower probability of mortality. The AI recommendation is to take the action with the highest value; in this case, it is to administer no vasopressors, along with a total fluid volume of between 500 and 960 mL over the time interval from 4 to 8 hours in Figure 1. All actions which are significantly lower in value than the AI recommendation are rejected, and the remaining actions are accepted, yielding the ranges of actions shown in Figure 1. Overall, clinician actions are deemed suboptimal by the AI with 99% confidence (Bonferroni corrected) in 69.9% of observed actions.

|                   | 1        | 2          | 3          | 4           | 5             | 6            |
|-------------------|----------|------------|------------|-------------|---------------|--------------|
| Fluid volume (mL) | [0, 100] | (100, 270] | (270, 500] | (500, 960]  | (960, ∞)      |              |
| Maximum VIS       | 0        | (0, 3.0]   | (3.0, 6.0] | (6.0, 10.0] | (10.0, 20.0]  | (20.0, ∞)    |

**Table 1:** Discretized cutoffs for total fluids and maximum vasopressor dosage over each 4-hour window. Each action is denoted as an ordered pair of fluid and vasopressor bins, yielding a space of 30 discrete actions.

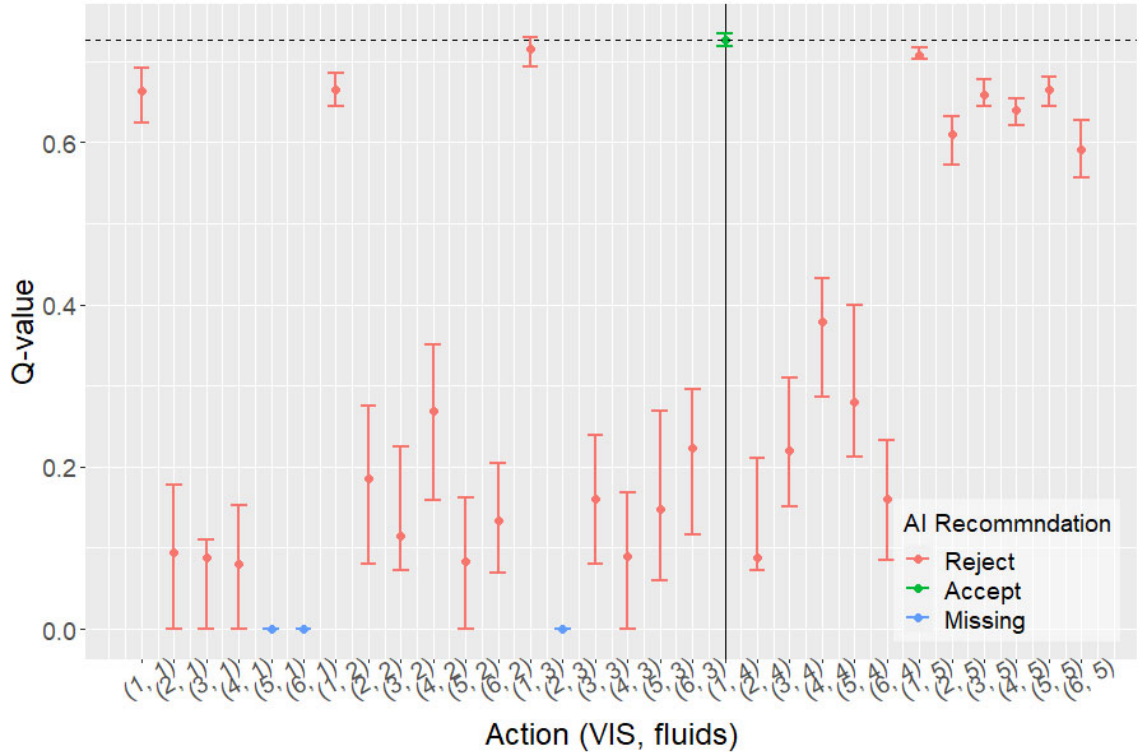

**Figure 2:** Confidence bounds of Q-values (action values) for state 48 estimated using bootstrap. The action with the highest value is recommended by the AI, and is denoted by a vertical line, and its value is denoted by the dashed horizontal line. Actions with value less than the AI recommended action with 99% confidence (Bonferroni correction, Wilcoxon rank-sum test) are indicated in red, and accepted actions are indicated in green.

The value of a policy is equal to the expected value of the policy-chosen actions over the distribution of starting states, which is equal to the dot product of the value of the AI chosen action in each state and the empirically observed probability of starting in each state in the data. We compare four policies (Figure 3A): the AI policy, the clinician policy (which takes actions with the same observed frequency as clinicians in the dataset), the random policy (which takes each action with equal probability), and the zero-intervention policy (which always administers 0 vasopressors, and the lowest bin of fluids). We computed the average mortality over twenty bins of action values (Q-values). Mortality is monotonically decreasing with respect to Q-value (Figure 3B), and thus a higher policy value corresponds to a lower expected mortality. The AI policy is constructed as the policy which maximizes Q-value given the state and action space formulation and observed transitions, and has a significantly higher expected value than all other policies (Wilcoxon rank-sum test, p<0.01, Bonferroni corrected). The value of the clinician and zero intervention policies are not significantly different from each other, but are higher than the value of the random policy, which yields the lowest value.

We can also visualize the approximate similarity between each pair of policies (Figure 3C, D). For each policy, we compute the recommended fluids and vasopressors for each patient over the course of their hospital stay, and compute the similarity for each pair of policies as the mean squared error between the time series of recommended actions. We can then compute the average similarity over all patients for each pair of policies, and use multi-dimensional scaling[37] to visualize these pair-wise similarities in two dimensions. We see that the AI fluid policy is most similar to the zero intervention policy, and more similar to the clinician policy than the random policy (Figure 3C). This is because the AI policy avoids large volumes of fluids in non-sepsis and sepsis without shock (Figure 4), whereas the random policy will pick those actions uniformly, and clinicians will choose those actions more frequently than the AI. We see the same pattern of similarity in the vasopressor policies (Figure 3D).

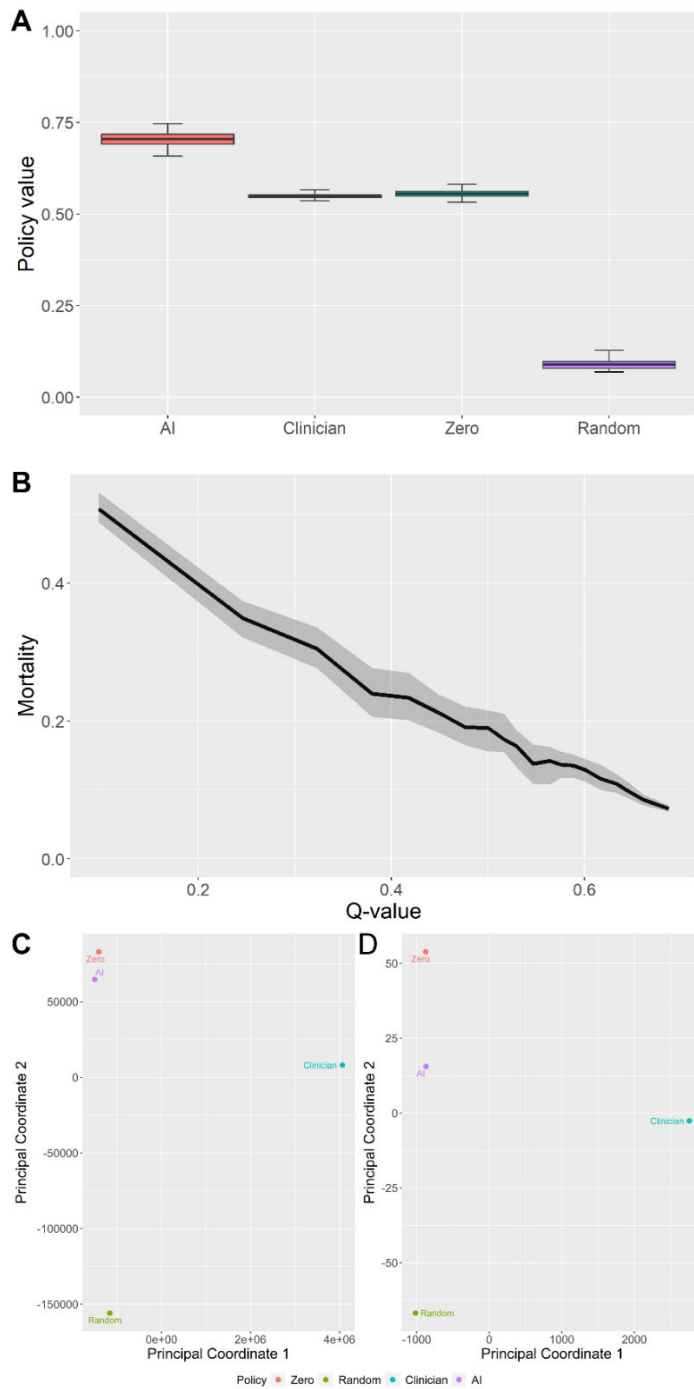

**Figure 3:** Estimated values for clinician, AI, random, zero policies (**A**). Relationship between AI valuation of observed clinician actions (Q-value) and mortality, with 95% confidence bound shaded (**B**). Policy similarity visualized in two dimensions (closer is more similar) for recommended fluids (**C**) and vasopressors (**D**).

Evaluating the AI recommendation at each of the observed states in the dataset yields marginal distributions for fluids and vasopressor recommendations (Figure 4). Both clinicians and the AI choose higher fluid volumes and dosages of vasopressors for patients in more severe clinical states (sepsis and septic shock) as compared to the less severe state (non-sepsis). The AI chooses the highest volumes of fluids more frequently in septic shock patients than clinicians, and chooses to administer intermediate volumes of fluids more commonly in sepsis and non-sepsis patients, whereas clinicians are distributed uniformly for sepsis patients, and prefer low or no fluids when patients are in non-sepsis. The most common dosage of vasopressors selected by both clinicians and the AI is zero. Vasopressors are administered by clinicians primarily in patients with septic shock, with a small proportion of patients in sepsis receiving vasopressors; AI recommendations also primarily administer vasopressors to septic shock patients, though the highest dosages of vasopressors are recommended at a lower frequency.

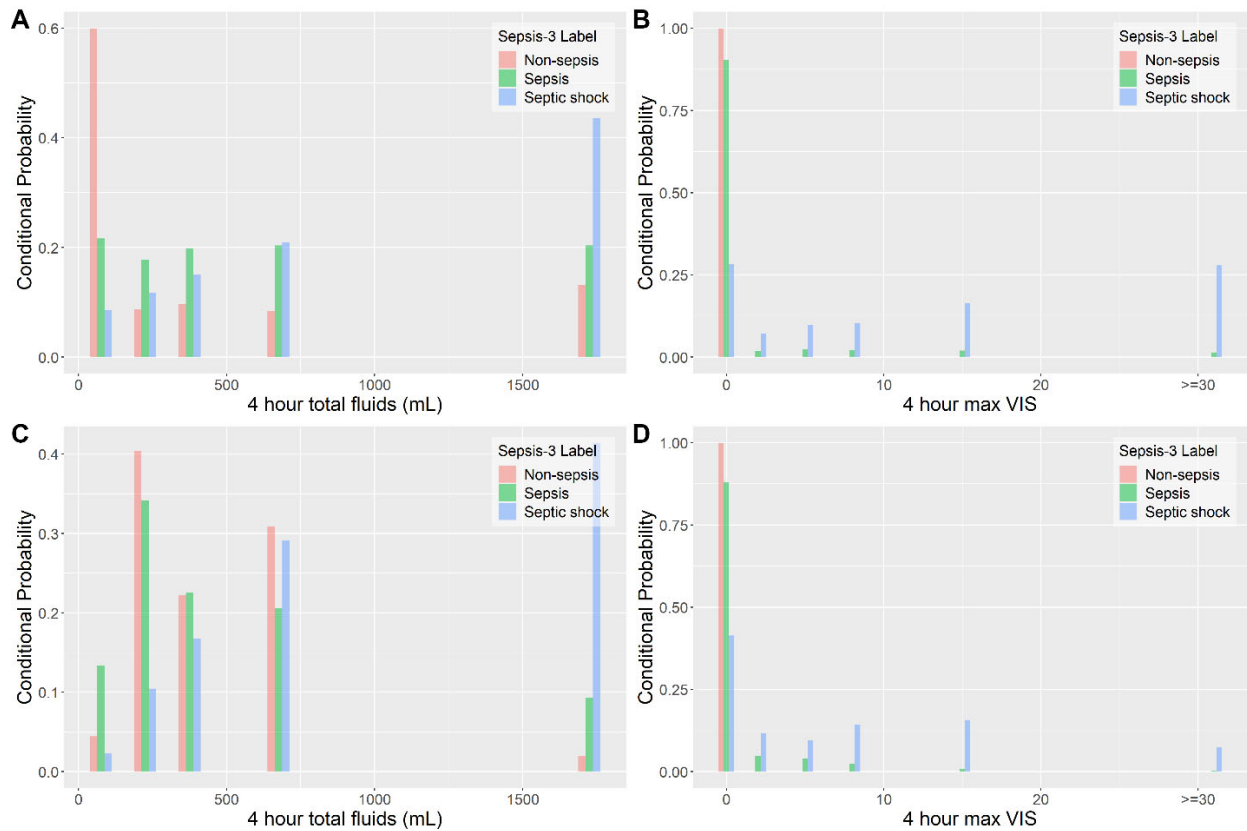

**Figure 4:** Distribution of actions (total fluid volume and maximum vasopressor dosage) by Sepsis-3 label for clinicians (**A**, **B**) and AI (**C**, **D**). Actions are binned into the discrete cutoffs specified in Table 1.

Different formulations of the reward function can be used to optimize for different treatment objectives. A terminal reward incurred upon patient discharge or mortality leads to optimization on overall patient mortality. We also examined the use of a mortality index to calculate a step-wise intermediate reward, used in conjunction with terminal reward. We fitted an XGBoost[38] model to the same physiological data used to compute patient state, producing a time-evolving prediction of the probability of mortality (Figure S2). Each transition that did not lead to discharge or mortality yielded a reward equal to the negative change in this mortality index over each 4-hour window: a decrease in the value of the mortality index would yield a positive reward value, and an increase in the value of the mortality index would yield a negative reward value. This allows optimization towards short-term resuscitation goals as well as overall patient outcome, where clinicians attempt to improve the hemodynamic status of the patient in the short term.

Changes in the value of the mortality index over each 4-hour window are used in determining intermediate reward, and thus the mortality index is used as a numeric representation of patient response to intervention in the short term based on changes in their observed physiological variables. The mortality index predicts mortality of the patient with a performance of 0.85 area under the receiver-operating curve (AUC). This XGBoost model-based index outperforms SOFA score in predicting patient prognosis (Figure 5). If a threshold is chosen according to the point on the receiver-operating (ROC) curve closest to the top left, this yields 79.2% accuracy, 73.1% sensitivity, 80.0% specificity, and 33.8% positive predictive value. Table S3 lists the most important features contributing to the value of this mortality index. Glasgow coma scale[39], blood urea nitrogen, bilirubin, and serum lactate concentration are the four most important features in the XGBoost model, and thus, improvement or deterioration in these features is captured by the mortality index.

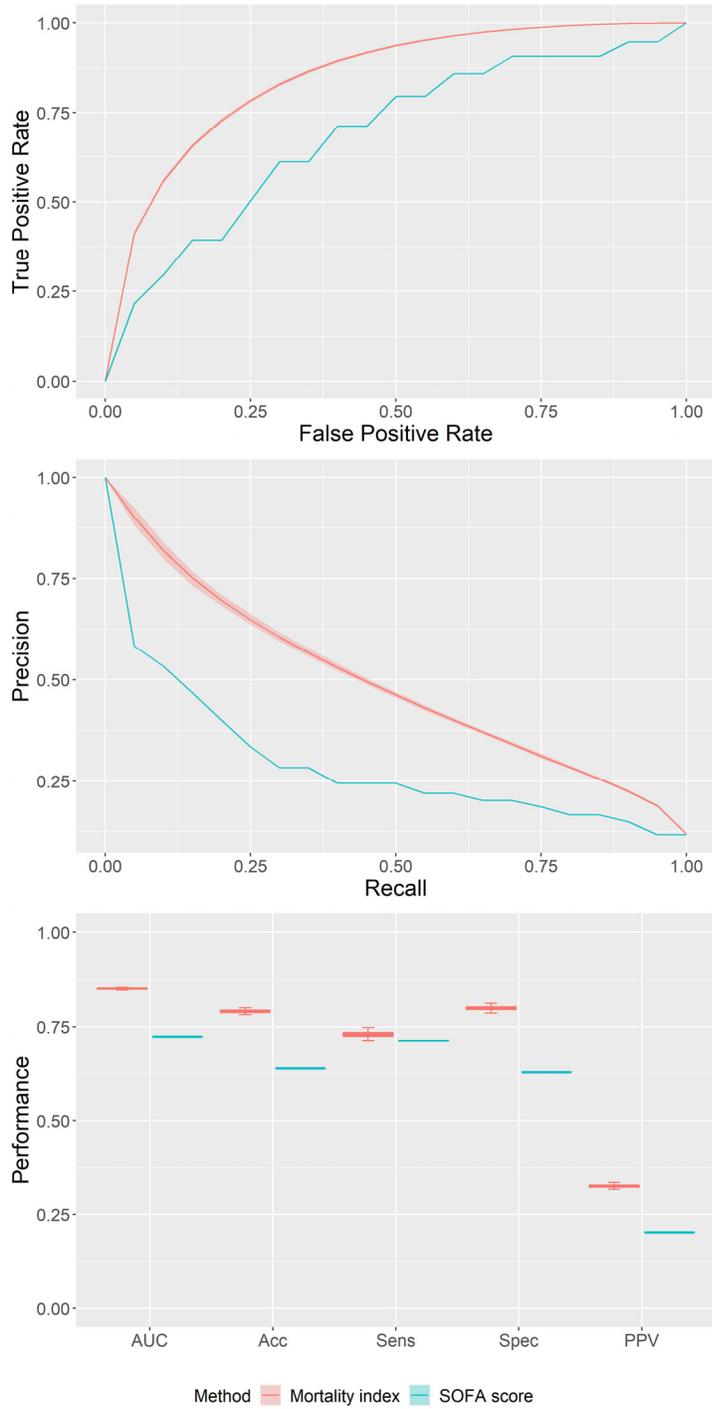

**Figure 5:** Mortality index performance: receiver operating characteristic curves with 95% confidence bounds shaded (**A**), precision-recall curves with 95% confidence bounds shaded (**B**), and performance criteria (**C**) for XGBoost mortality index and SOFA score.

This leads to AI recommendations which are in greater agreement with clinician actions (64.0% of observed actions are rejected with intermediate reward vs 69.9% with only terminal reward), and a 13.6% decrease in mean fluid usage recommended by the AI, from 485 mL over 4 hours to 419 mL (Figure 6). The introduction of intermediate reward results in a reduction of recommended fluid administration in non-sepsis patients (Figure 6B), a reduction of the recommendation of the highest quantities of fluids in sepsis and septic shock patients (Figure 6D, F), a slight increase in the recommendation of higher dosages of vasopressors in sepsis patients without septic shock (Figure 6C), but an overall decrease in the frequency of vasopressor usage in sepsis and septic shock patients (Figure 6C, E).

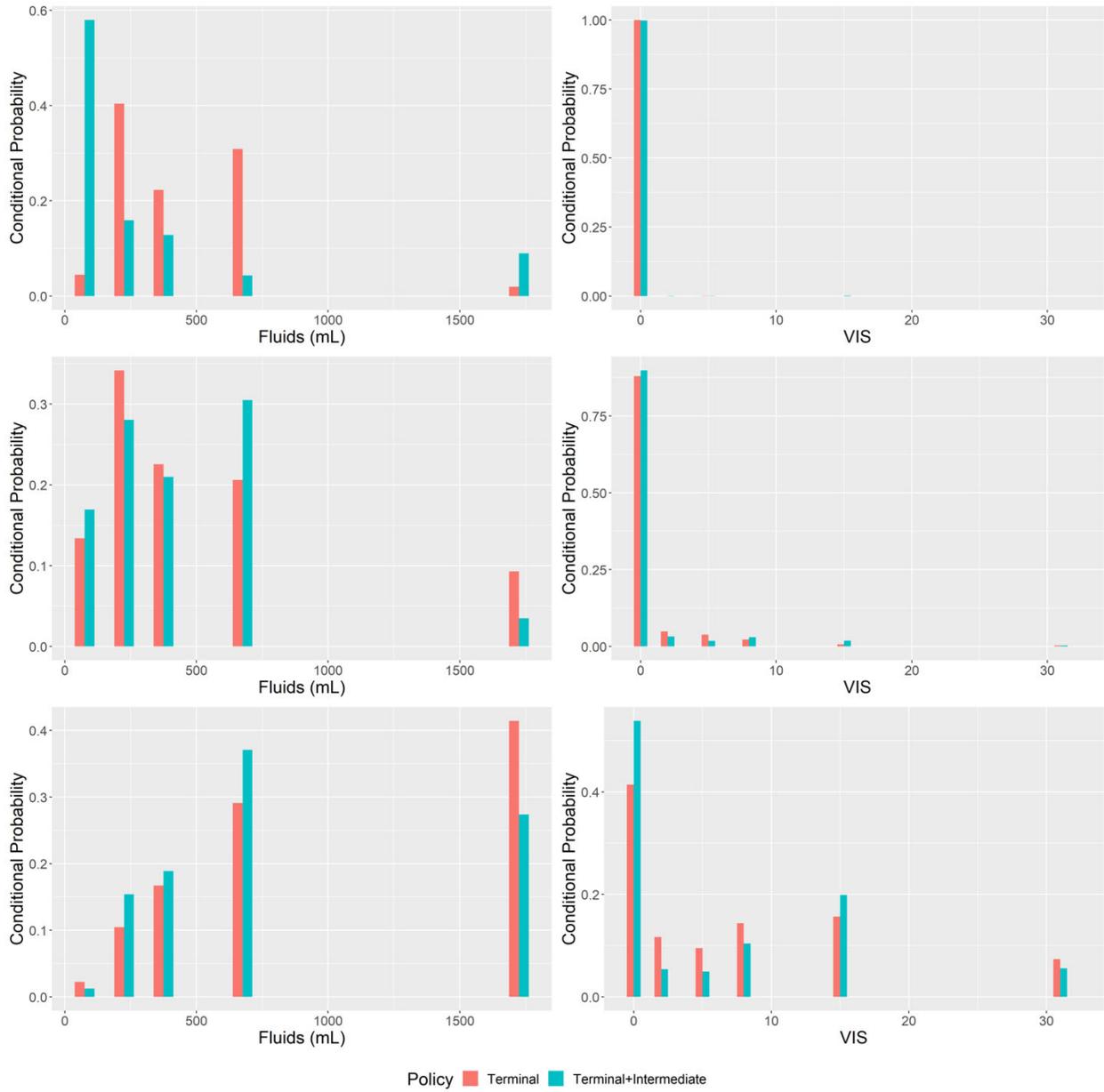

**Figure 6:** Comparison of frequency of recommended actions before and after the addition of intermediate reward based on change in mortality index for vasopressors (**A**, **C**, **E**) and fluids (**B**, **D**, **F**) in non-sepsis (**A**, **B**), sepsis (**C**, **D**), and septic shock (**E**, **F**).

Dimensionality reduction of physiological data with CCA prior to k-means clustering yields clusters which are more homogenous with respect to clinical states (Figure 7), in interventions received, and in mortality outcome. That is, physiological features from time points where patients are in septic shock and when patients are not in septic shock are more likely to cluster into separate states after CCA: there is greater overlap of the distributions in Figure 7A than in 7C. Moreover, when data from septic shock patients is clustered into the same states as data from patients not in septic shock, they are more similar in treatments received after CCA: because in our model formulation, actions taken are conditioned solely on state, it is undesirable for there to be distinguishable groups within each state for whom clinicians choose different actions. This indicates that the clinicians are making their treatment decisions based upon information that is not captured by the state-space. Patients in this cohort can be separated into groups by clinical state label. Therefore, we examine septic shock and non-shock, comprised of non-sepsis and sepsis without septic shock.

Figure 7B and 7D illustrate the distribution of vasopressor dosages administered in patients in the 15 states with the highest prevalence of septic shock; in Figure 7B, the state-space is defined by clustering on normalized features from structured EHR data and treatment history, whereas in Figure 7D, we apply CCA for dimensionality reduction (note that analyses shown in other figures used CCA before discretization of the state-space). There is a much greater difference in the median dosage of vasopressors given to septic shock and non-shock patients in Figure 7B, where CCA is not used than in Figure 7D, where clustering is performed after CCA. Because clinicians give similar vasopressor dosages to all patients within each state when CCA is applied prior to clustering, we believe that the resulting state-space captures more of the information which is relevant when choosing a course of intervention in sepsis patients.

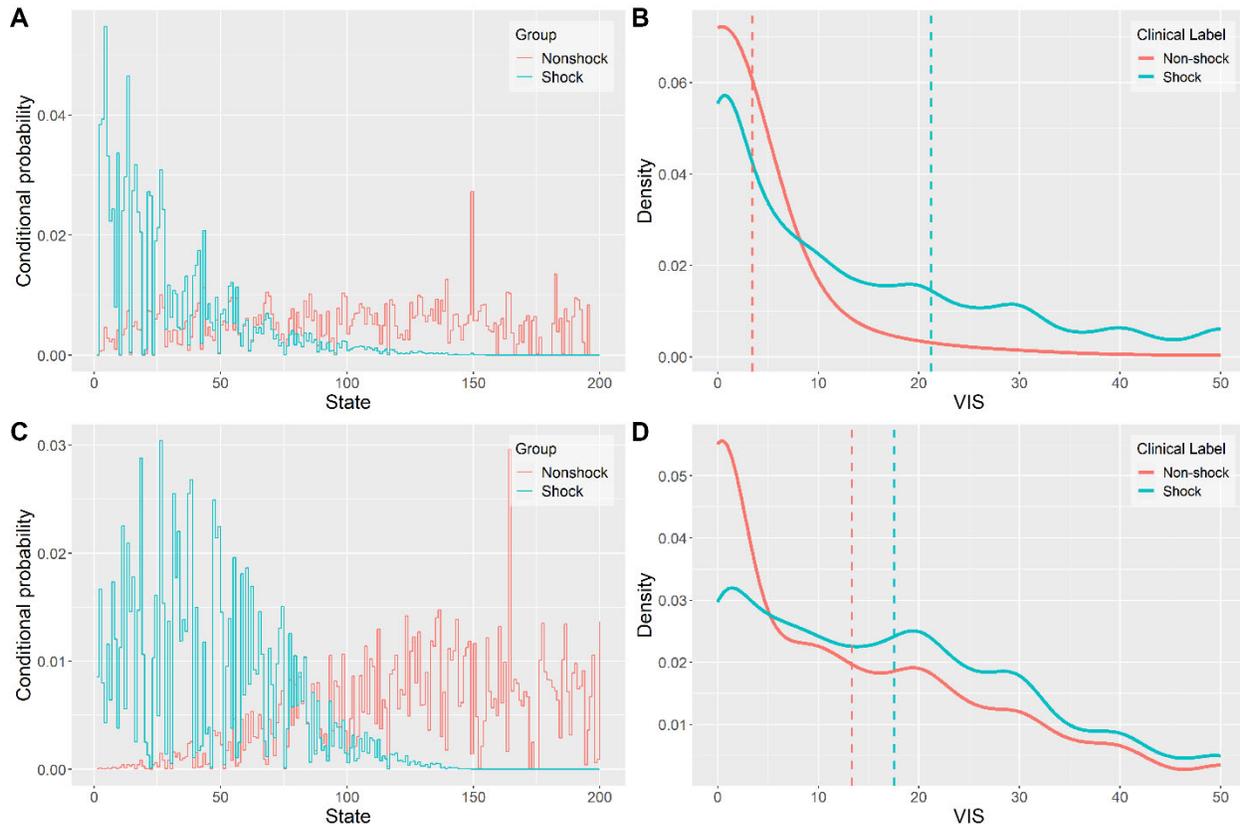

**Figure 7:** Probability of occupying each state conditioned on clinical state label (shock vs non-shock), with states defined by k-means clustering on physiological data (**A**) and physiological data after dimensionality reduction with CCA (**C**). For the 15 states with the highest prevalence of septic shock, the kernel density estimates of the conditional distributions of vasopressors received for patients in shock and not in septic shock (**B**, **D**). Dotted vertical lines indicate the median of each distribution.

**Discussion:**

*Overview*

We use reinforcement learning to compute treatment policies in sepsis patients. A main result is that we are able to compute ranges of actions which are accepted, rejected, or unobserved, rather than prescribing a single action. By providing ranges of actions which are comparable in value to the AI recommendation, we provide the clinician with more substantive information.

We examine different reward formulations representing different treatment objectives, contrasting the terminal objective of minimizing sepsis patient mortality against the short-term goal of correcting an immediate hemodynamic disturbance. This short-term treatment objective more closely resembles the goal of current sepsis treatment guidelines, and yields AI policies which are more similar to observed clinician actions.

Finally, we underscore the importance of intra-state homogeneity when formulating the underlying Markov decision process model, and reveal that the existence of heterogeneous states yields AI policies which are biased towards inaction. We ameliorate this problem through learning a linear subspace of the EHR data which maximizes correlation with outcomes and interventions received using CCA: we learn weightings of the most relevant features, so that clustering yields states which are more homogenous in important features.

*Markov Decision Processes*

Reinforcement learning methods fundamentally represent the problems which they attempt to solve as Markov decision processes (MDP), which are discrete time Markov chains with transition probabilities conditioned on actions taken at each discrete time step. The main assumption in this formulation is that the Markov property holds for each state transition, and that the only factors which determine transition probabilities are the observed state of the patient, and the action taken by the clinician.

These assumptions don't precisely hold in practice: discretized states and actions will not be perfectly homogenous. Patients with varying degrees of disease severity, and different responsiveness to treatment, may be combined into a single state. This can be problematic, as healthy patients tend to receive lower levels of intervention than severely ill patients, and yet have good outcomes. If the two groups are combined into the same discrete states, then the AI policy will choose a lower level of intervention for all patients in that state.

However, there are theoretical advantages to formulating the clinical problem of hemodynamic optimization of sepsis patients as a discrete state, discrete action space Markov decision process. Most importantly, from the Banach fixed point theorem[24,40], we have guaranteed convergence of the Bellman equation (see *Materials and Methods, Q-learning*) to a unique optimal solution, i.e. a unique policy which maximizes expected value, given the distribution of initial states observed in the data.

Reinforcement learning methods with continuous state spaces which use deep neural networks to approximate value functions lose this theoretical guarantee, though deep reinforcement learning methods still generally offer good performance in practice. Moreover, when used for offline reinforcement learning, deep reinforcement learning methods encounter the problem of distribution shift[29], where neural network models tend to overestimate the value of actions which are infrequently observed or unobserved in the training data. Nonetheless, other groups have applied such methods to generate treatment policies for sepsis patients[27,41]. There are several recent studies on algorithmic approaches which attempt to address the challenges of offline deep reinforcement learning[31,32,42].

In the online case, where additional training can be conducted using learned policies, new observations can be used to correct errors that arise in training. However, when applying reinforcement learning to problems in healthcare, it is non-trivial to allow AI to dictate the treatment of patients. Without online interaction, it is difficult to assess the true performance of learned policies, or to correct errors in them. But the value of reinforcement learning treatment policies is not necessarily to automate the treatment of patients so much as it is to try to learn the best treatment policies. We derive value through examining the resulting AI treatment policies against observed clinician practice and current best practices. Differences in AI recommendations and observed actions allow us to form hypotheses about treatment of sepsis and septic shock patients which can be further studied and evaluated; the gold standard for any treatment guideline would be to demonstrate improved outcome in a randomized control trial, like those conducted for early goal-directed therapy[43-45].

From a methodological perspective, we can at least identify several likely causes of error in reinforcement learning for treatment policies in sepsis patients: the Markov decision process which models the treatment problem must reflect reality as closely as possible, the data on clinician interventions must be complete (as missing interventions would result in the wrong actions in the training data), and the data must be sufficient in quantity that the value of each action can be estimated. The other benefit of online training in this regard is that sometimes, the AI may recommend actions which are rarely taken by the clinician. If these actions are erroneously overvalued, then exploration using the learned policy will yield more observations which can be used to correct these values. However, when only offline learning is possible, then observations of these actions must already be plentiful in the training data for an accurate evaluation to be made.

*State Space*

K-means clustering uses Euclidean distance between data points to assign them to clusters, and thus is subject to the curse of dimensionality. Therefore, dimensionality reduction alone will tend to improve the quality of clustering. Moreover, similarity in certain features is more important than similarities in others. Using CCA as a dimensionality reduction method prior to clustering addresses both issues simultaneously, by learning linear weightings of features that maximize correlation with relevant outcomes such as clinical labels and mortality, and interventions. This results in clusters which are more homogenous in relevant features, which is a necessary property of a discrete state space for reinforcement learning to generate sensible policies.

Lack of resolution resulting in heterogenous states leads to reinforcement learning policies which are biased against higher volumes of fluids and higher dosages of vasopressors. This is because of the confounding relationship between patient mortality and level of treatment administered. Sicker patients will have a higher mortality rate, yet will also receive more aggressive intervention. However, if these patients are clustered into the same states as patients who are not as sick, who receive lower levels of intervention yet have a lower mortality rate, the two cohorts will be treated as indistinguishable by the underlying Markov decision process, and reinforcement learning will choose the lower level of intervention for all patients within those states. This problem is present not only in the treatment of sepsis and septic shock, but across problems in healthcare where this confounding relationship between intervention and mortality exists. Our methodology for mitigating this bias through subspace learning on the state space may therefore be applicable across applications of reinforcement learning to problems in healthcare.

Furthermore, patients with identical physiological variables may be in very different condition, if one has been treated, and the other has not. We capture past intervention history in patient state by including the past 12 hours of vasopressors and fluid volume administered.

*Action Space*

We discretized the action space according to the range of clinician actions observed in the dataset. While reinforcement learning is able to learn strategies not observed in the dataset, it is unable to evaluate actions which it has not observed. For example, in non-sepsis patients, clinicians do not administer vasopressors in most cases (Figure 4B), and so it is impossible for the AI to recommend vasopressors in those states. The decision for a clinician to initiate vasopressors is clinically significant, and so a value of 0 vasopressors administered in the past 4 hours is a separate bin from the non-zero quintiles, whereas for fluids, a value of 0 is included in the lowest quintile.

*Confidence Bounds*

Bootstrap is commonly used in estimating uncertainty in reinforcement learning, but these estimates are usually coarse, sometimes using as few as five samples, when used in deep reinforcement learning, as each iteration requires fitting a new model to the sampled data[33,46,47]. Our discrete state MDP only requires a few minutes to reach convergence, as there are only 6000 parameters (the product of the sizes of the state and action spaces) compared to the tens or hundreds of millions which are sometimes present in deep learning models.

The ability to identify actions whose values are uncertain to the AI may be valuable for clinicians when taking into account treatment recommendations. It allows us to identify a range of actions whose values are not significantly different from that of the AI recommendation. Estimates of uncertainty provide clinicians with more information, and more discretion than a simple black box recommendation. If an action with high uncertainty in its value is chosen by the AI, then many other actions may have values within its confidence bound, giving a wider range of options to the clinician in cases where there may be doubt in the AI recommendation. The main source of uncertainty captured by bootstrap is the infrequent observation of certain actions in any given state. If only a few observations are present in the data, with disparate outcomes, then the value of that action may change greatly across each iteration, resulting in high uncertainty. If outcomes are more homogenous across the observations, then there will be lower variance in the action value across bootstrap iterations.

*AI Policy*

Broadly, there are some comparisons which can be made between the learned AI policy and current clinical practice which can assure us that the recommendations made are at least roughly sensible. Generally, we expect that greater levels of intervention to sicker patients, which we observe in the clinician actions: septic shock patients receive the greatest amounts of fluids, and sepsis patients receive greater fluid volumes than non-sepsis patients (Figure 4). Vasopressors are not administered in non-sepsis patients, and only in a small fraction of sepsis patients, whereas most septic shock patients received vasopressors in the past 4 hours. We observe this same pattern in the AI recommendations, though there are differences in the frequencies that specific actions are recommended.

There is a greater discrepancy between clinician actions and AI recommendations in fluid volume than vasopressor dosages. However, this is at least partially due to the fact that most patients do not receive vasopressors, and if no vasopressor usage is observed in any given state, then the AI cannot recommend vasopressor usage. For this reason, there are fewer opportunities for the AI to disagree with clinicians on vasopressor dosage in the observed data, and thus, naturally, fewer disagreements between the AI and clinicians.

*Intermediate Reward*

Different reward structures reflect different objectives for treatment with vasopressors and fluids in sepsis patients. We used change in mortality index in order to represent the short-term goal of a clinician to improve the physiological state of a sepsis or septic shock patient. The moderately high performance achieved by the mortality index in predicting patient mortality is evidence that it functions as a reasonable indicator of patient state. The greater similarity of the AI policy to clinician actions after the addition of intermediate reward seems to indicate that this may be somewhat similar to the actual object of clinicians when administering fluids and vasopressors. The biggest difference in AI policy after the addition of intermediate reward is that fluid administration is reduced in non-sepsis patients (Figure 6). This may be a sensible change, as patients not in sepsis may no longer require further administration of fluid volume.

*Limitations*

The primary limitation of reinforcement learning in healthcare applications in general is the inability to train and test in an online setting. This prevents the discovery and correction of many types of errors arising from biases in the dataset or missing information on the actions taken by clinicians, and also makes it difficult to evaluate the true performance of a policy. Retrospective observational studies in general are constrained in their ability to validate their conclusions; however, there may be a considerably greater obstacles to prospective usage of AI that directly provide treatment recommendations. While bootstrap estimates of uncertainty are able to capture variance in Q-values when the data is drawn from the same distribution as the training data, it is unable to detect biases or coincidental associations due to sampling variance in the training dataset itself. In other reinforcement learning applications, errors of this nature would be corrected by additional online training.

Though we avoid the problem of distribution shift in offline deep reinforcement learning by using a discrete state space, this introduces limitations in state resolution, and states which are to some degree heterogenous. We improve the homogeneity of our discrete states through dimensionality reduction with CCA, but some heterogeneity still remains (Figure 7). Outliers will still be clustered with whichever state is closest, even if the majority of data points are very dissimilar. However, in outlier regions of sparse data, deep reinforcement learning methods will also find themselves out of distribution, and unlikely to perform well either. Because no further online interaction is possible, it is necessary that the training data be well-populated with sufficient examples of different clinician actions. K-means clustering produces clusters of roughly equivalent size; however, some state-action combinations will only be sparsely populated. Septic shock patients comprise only a small portion of the dataset, yet most vasopressor usage occurs in patients with septic shock.

Because in formulating the problem of intervention as a Markov decision process, we assume that the only factors determining transition probabilities, and thus outcomes, are patient state and clinician actions, it is important that this information is as complete and high-quality as possible. If any data on interventions is missing, then clinician action in that window will erroneously be marked as a lower level of intervention, shifting estimates of the value of each action. Similarly, missing features would also impact the state assignment of a patient. It is likely if the clinician has more or better information than the AI, that the clinician will yield better outcomes despite having a lower policy value, as the clinician policy assumes that actions are randomly sampled according to the frequencies observed for each action by the clinicians in each state. If the clinician has greater knowledge of the patient's state than the AI, then this assumption does not hold, and the clinician policy does not model the actual behavior of clinicians very well.

**Materials and Methods:**

*Data Extraction and Processing*

We extract data using the same procedure as in our previously described methods[11]. The MIMIC-III database (version 1.4)[34] contains data from 52,501 adult ICU stays from patients admitted to Beth Israel Deaconess Medical Center from 2001 to 2012. Adult patients labeled as having sepsis at least once during their ICU stays were retained for study, and the rest excluded. The majority of data entries are comprised of timestamp-value pairs. A complete table of all item ids corresponding to each feature used in this study is included in Table S2. From the queried data, we compute values over time for each feature over the course of each ICU stay, carry forward the most recent known observation, and impute with the population mean value where no previous observation is available.

*Clinical Labels*

The Third International Consensus Definitions for Sepsis and Septic Shock[5] were applied to patient EHR data at each time where there were observations. Suspected infection was determined as recommended by Seymour et al., using concomitant orders for antibiotics and blood cultures[48]. Sepsis patients are those with suspected infection and a Sequential Organ Failure Assessment (SOFA)[49] score of 2 or higher. Septic shock patients are those who have sepsis, have received adequate fluid resuscitation as determined according to the Surviving Sepsis Campaign guidelines[6,10], have received vasopressors, and have a serum lactate of 2 mmol/L or greater.

*Action Space*

Discrete cutoffs for actions are determined by splitting the space of non-zero total fluid volumes administered into quintiles. Vasopressor dosages are represented by vasoactive-inotropic score[35].

The absence of vasopressor administration is a separate bin, resulting in a total of 6 vasopressor bins, and 5 fluid bins. The action space is defined by the cartesian product of fluids and vasopressors, i.e. by the combination of a fluid and vasopressor action, yielding 30 discrete actions.

*State Space*

We learn linear weightings of features that maximize correlation with outcomes and interventions received using canonical correlation analysis[36] between the physiological EHR variables with treatment history over the past 12 hours, and the observed clinician action, patient's terminal mortality outcome, and clinical state label according to the Sepsis-3 criteria. Using these weightings, patient data is transformed into a new space of canonical correlates, where each feature is a linear combination of physiological data and past intervention history that is maximally correlated with outcomes and interventions received.

The state space is defined using k-means clustering on the top 5 canonical correlates. We use the elbow heuristic on Akaike information criterion (AIC) to select the number of clusters, resulting in 200 discrete states defined by clustering on physiological data and past intervention history. For the purposes of the underlying Markov decision process, there are also two absorbing states of discharge and death, though no actions exist for these states. Every episode, or sequence of observed patient states, clinician actions, and step-wise rewards for a given ICU admission terminates in either discharge or death.

*Reward Function*

Terminal reward occurs upon transition into the absorbing states of discharge or death. We assign a value of +1 to discharge, and -1 to death. Intermediate reward is defined as the negative change in predicted risk of mortality over each 4-hour window, where a decrease in mortality risk is rewarded, and an increase in mortality risk is penalized. Risk of mortality is predicted using an XGBoost[38] model with the same features which are used to compute the state space. Each data point is classified as from a patient who dies after hospital admission, or is discharged, and the risk score is the resulting predicted probability of mortality.

*Q-learning*

Figure 8 illustrates one transition in the Markov decision process we used to model sepsis interventions, which is a discrete time Markov chain where actions must be taken at each time step. State transition probabilities are determined by both the starting state, as well as the action taken at time t. Rewards for each time step depend upon the starting state ($s_1$ through $s_{200}$), the action taken ($a_1$ through $a_{30}$, Table 1, Figure 2), as well as the state into which the transition occured ($s_1$ through $s_{200}$, plus the absorbing states of death and discharge).

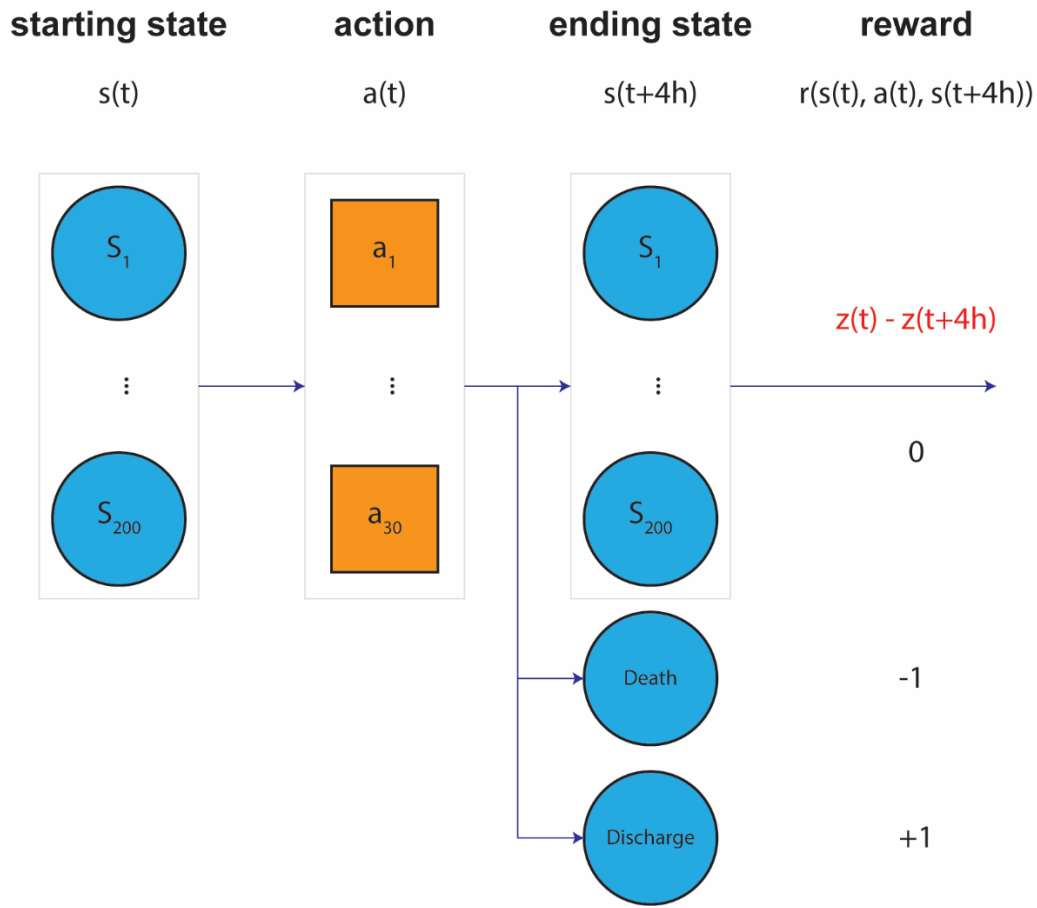

**Figure 8:** Illustration of a single transition in the Markov decision process (MDP) used to model sepsis interventions. There is a starting state, s(t), an action taken, a(t), and an ending state, s(t+4h). Transition probabilities between states depend upon starting state and the action taken. There are two absorbing states, death and discharge, upon which an episode ends, and a terminal reward is incurred. Otherwise, another transition occurs at the next time step. If intermediate reward is used, then transitions into non-absorbing states incur a reward (indicated in red) equal to the negative change over the 4-hour window in z(t), the mortality index, which is the predicted probability of mortality. Otherwise, reward for these transitions is 0.

Q-learning is an algorithm for learning the expected reward value of each action in a Markov decision process. The Bellman equation for a Markov decision process[24] is given by Equation 1, specifying a recursion for the expected value of the total reward after taking each action in a given state, following a pre-specified policy. When the policy followed is specified as the optimal policy which maximizes the expected reward, this is not directly solvable as a linear system of equations, as the maximum function over each action is non-linear. Nonetheless, we are guaranteed convergence to a unique optimal solution by the Banach fixed point theorem, so long as we have a contractive mapping.

$$q^\pi(s, a) = \sum_{s'} p(s'|s, a) \left( r(s, a, s') + \gamma \sum_{a'} \pi(s', a') q(s', a') \right)$$

**Equation 1:** The Bellman equation for a Markov decision process. $q^\pi(s, a)$ denotes the expected value for a given state, s, and action, a, when taking actions in all subsequent time steps according to a policy π, which is a probability distribution over the discrete action space for each state, s. It is equal to the reward incurred from the current observed action, plus the expected value over all possible destination states.

For performance, we implement Q-learning as an optimization in Tensorflow, solving the system as a loss minimization using batch gradient descent.

*Estimating Uncertainty in Action Valuations*

Estimating uncertainty in action valuations is done using bootstrap. Patients are sampled with replacement from the training data, and a bootstrapped dataset containing their observed transitions is used to learn a new value function. The resulting distribution of action valuations across all models is then an estimate of the distribution of values for each action, conditioned on state. We run 100 bootstrap iterations for each different formulation of Markov decision process. Acceptance or rejection of actions is evaluated using the one-sided Wilcoxon rank-sum test[50] against the AI recommendation of the action with the greatest expected value for reward, with Bonferroni correction for multiple hypothesis testing.

**Data Availability:**

Data analyzed in this study are publicly available from the MIMIC-III database (https://mimic.physionet.org), version 1.4.

**Acknowledgements:**

This work was funded by NSF EECS 1609038 and NIH UL1 TR001079. We would like to thank Dr. Nauder Faraday and Adam Sapirstein for valuable discussion. We would also like to thank Eric Bridgeford for statistical and mathematical consultation.

**Competing Interests:**

The authors declare that there are no competing interests.

**Author Contributions:**

RL contributed to project design, developed and implemented methods, performed computations, analyzed and interpreted results, and drafted the original manuscript. JLG and JB contributed to data analysis, interpretation, and final drafting of the manuscript. JCF and MMB contributed to data interpretation and final drafting of the manuscript. RLW contributed to project design and direction, and final drafting of the manuscript.

**References:**

1. Liu, V. *et al.* Hospital Deaths in Patients With Sepsis From 2 Independent Cohorts. *Jama* **312**, 90, doi:10.1001/jama.2014.5804 (2014).
2. Torio, C. M. & Moore, B. J. in *Healthcare Cost and Utilization Project (HCUP) Statistical Briefs* (2016).
3. Hume, P. S. *et al.* Trends in "usual care" for septic shock. *Infection Control & Hospital Epidemiology* **39**, 1125-1126, doi:10.1017/ice.2018.154 (2018).
4. Stevenson, E. K., Rubenstein, A. R., Radin, G. T., Wiener, R. S. & Walkey, A. J. Two Decades of Mortality Trends Among Patients With Severe Sepsis. *Critical Care Medicine* **42**, 625-631, doi:10.1097/ccm.0000000000000026 (2014).
5. Singer, M. *et al.* The Third International Consensus Definitions for Sepsis and Septic Shock (Sepsis-3). *Jama* **315**, 801, doi:10.1001/jama.2016.0287 (2016).
6. Rhodes, A. *et al.* Surviving Sepsis Campaign: International Guidelines for Management of Sepsis and Septic Shock: 2016. *Intensive Care Medicine* **43**, 304-377, doi:10.1007/s00134-017-4683-6 (2017).
7. Kumar, A. *et al.* Duration of hypotension before initiation of effective antimicrobial therapy is the critical determinant of survival in human septic shock*. *Critical Care Medicine* **34**, 1589-1596, doi:10.1097/01.ccm.0000217961.75225.e9 (2006).
8. Martin-loeches, I., Levy, M. & Artigas, A. Management of severe sepsis: advances, challenges, and current status. *Drug Design, Development and Therapy*, 2079, doi:10.2147/dddt.s78757 (2015).
9. Ferrer, R. *et al.* Empiric Antibiotic Treatment Reduces Mortality in Severe Sepsis and Septic Shock From the First Hour. *Critical Care Medicine* **42**, 1749-1755, doi:10.1097/ccm.0000000000000330 (2014).
10. Levy, M. M., Evans, L. E. & Rhodes, A. The Surviving Sepsis Campaign Bundle: 2018 Update. *Critical Care Medicine* **46**, 997-1000, doi:10.1097/ccm.0000000000003119 (2018).
11. Liu, R. *et al.* Data-driven discovery of a novel sepsis pre-shock state predicts impending septic shock in the ICU. *Scientific Reports* **9**, doi:10.1038/s41598-019-42637-5 (2019).
12. Henry, K. E., Hager, D. N., Pronovost, P. J. & Saria, S. A targeted real-time early warning score (TREWScore) for septic shock. *Science Translational Medicine* **7**, 299ra122-299ra122, doi:10.1126/scitranslmed.aab3719 (2015).
13. Mao, Q. *et al.* Multicentre validation of a sepsis prediction algorithm using only vital sign data in the emergency department, general ward and ICU. *BMJ Open* **8**, e017833, doi:10.1136/bmjopen-2017-017833 (2018).
14. Lauritsen, S. M. *et al.* Explainable artificial intelligence model to predict acute critical illness from electronic health records. *Nature Communications* **11**, doi:10.1038/s41467-020-17431-x (2020).
15. Liu, R. *et al.* Early prediction of impending septic shock in children using age-adjusted Sepsis-3 criteria. *medRxiv*, 2020.2011.2030.20241430, doi:10.1101/2020.11.30.20241430 (2020).
16. Kelm, D. J. *et al.* Fluid Overload in Patients With Severe Sepsis and Septic Shock Treated With Early Goal-Directed Therapy Is Associated With Increased Acute Need for Fluid-Related Medical Interventions and Hospital Death. *Shock* **43**, 68-73, doi:10.1097/shk.0000000000000268 (2015).

17    Durairaj, L. & Schmidt, G. A. Fluid Therapy in Resuscitated Sepsis. *Chest* **133**, 252-263, doi:10.1378/chest.07-1496 (2008).
18    Dünser, M. W. *et al.* Association of arterial blood pressure and vasopressor load with septic shock mortality: a post hoc analysis of a multicenter trial. *Critical Care* **13**, R181, doi:10.1186/cc8167 (2009).
19    Schmittinger, C. A. *et al.* Adverse cardiac events during catecholamine vasopressor therapy: a prospective observational study. *Intensive Care Medicine* **38**, 950-958, doi:10.1007/s00134-012-2531-2 (2012).
20    Pollack, L. A. & Srinivasan, A. Core elements of hospital antibiotic stewardship programs from the Centers for Disease Control and Prevention. *Clinical Infectious Diseases* **59**, S97-S100 (2014).
21    Minderhoud, T. C. *et al.* Microbiological outcomes and antibiotic overuse in Emergency Department patients with suspected sepsis. *Neth J Med* **75**, 196-203 (2017).
22    Seymour, C. W. *et al.* Derivation, Validation, and Potential Treatment Implications of Novel Clinical Phenotypes for Sepsis. *Jama*, doi:10.1001/jama.2019.5791 (2019).
23    Liu, R., Greenstein, J. L., Fackler, J. C., Bembea, M. M. & Winslow, R. L. Spectral clustering of risk score trajectories stratifies sepsis patients by clinical outcome and interventions received. *eLife* **9**, doi:10.7554/eLife.58142 (2020).
24    Sutton, R. S. & Barto, A. G. *Reinforcement learning: An introduction*. (MIT press, 2018).
25    Komorowski, M., Celi, L. A., Badawi, O., Gordon, A. C. & Faisal, A. A. The Artificial Intelligence Clinician learns optimal treatment strategies for sepsis in intensive care. *Nature Medicine* **24**, 1716-1720, doi:10.1038/s41591-018-0213-5 (2018).
26    Roggeveen, L. *et al.* Transatlantic transferability of a new reinforcement learning model for optimizing haemodynamic treatment for critically ill patients with sepsis. *Artificial Intelligence in Medicine* **112**, 102003 (2021).
27    Peng, X. *et al.* Improving Sepsis Treatment Strategies by Combining Deep and Kernel-Based Reinforcement Learning. *AMIA Annu Symp Proc* **2018**, 887-896 (2018).
28    Yu, C., Ren, G. & Liu, J. in *2019 IEEE International Conference on Healthcare Informatics (ICHI)*.   1-3 (IEEE).
29    Levine, S., Kumar, A., Tucker, G. & Fu, J. Offline reinforcement learning: Tutorial, review, and perspectives on open problems. *arXiv preprint arXiv:2005.01643* (2020).
30    Liu, S. *et al.* Reinforcement Learning for Clinical Decision Support in Critical Care: Comprehensive Review. *Journal of Medical Internet Research* **22**, e18477, doi:10.2196/18477 (2020).
31    Yu, T. *et al.* Mopo: Model-based offline policy optimization. *arXiv preprint arXiv:2005.13239* (2020).
32    Kumar, A., Zhou, A., Tucker, G. & Levine, S. Conservative q-learning for offline reinforcement learning. *arXiv preprint arXiv:2006.04779* (2020).
33    Kumar, A., Fu, J., Tucker, G. & Levine, S. Stabilizing off-policy q-learning via bootstrapping error reduction. *arXiv preprint arXiv:1906.00949* (2019).
34    Johnson, A. E. W. *et al.* MIMIC-III, a freely accessible critical care database. *Scientific Data* **3**, 160035, doi:10.1038/sdata.2016.35 (2016).
35    Gaies, M. G. *et al.* Vasoactive–inotropic score as a predictor of morbidity and mortality in infants after cardiopulmonary bypass. *Pediatric Critical Care Medicine* **11**, 234-238 (2010).

36      Hotelling, H. Relations between Two Sets of Variates. *Biometrika* **28**, 321-377, doi:10.1093/biomet/28.3-4.321 (1936).

37      Friedman, J., Hastie, T. & Tibshirani, R. *The elements of statistical learning*. Vol. 1 (Springer series in statistics New York, 2001).

38      Chen, T. & Guestrin, C. in *Proceedings of the 22nd acm sigkdd international conference on knowledge discovery and data mining.* 785-794 (ACM).

39      Teasdale, G. & Jennett, B. Assessment of coma and impaired consciousness. A practical scale. *Lancet* **2**, 81-84 (1974).

40      Vershynin, R. *High-dimensional probability: An introduction with applications in data science*. Vol. 47 (Cambridge university press, 2018).

41      Li, L., Komorowski, M. & Faisal, A. A. Optimizing Sequential Medical Treatments with Auto-Encoding Heuristic Search in POMDPs. *arXiv preprint arXiv:1905.07465* (2019).

42      Agarwal, R., Schuurmans, D. & Norouzi, M. in *International Conference on Machine Learning.* 104-114 (PMLR).

43      A Randomized Trial of Protocol-Based Care for Early Septic Shock. *New England Journal of Medicine* **370**, 1683-1693, doi:10.1056/NEJMoa1401602 (2014).

44      Ding, X.-F. *et al.* Early goal-directed and lactate-guided therapy in adult patients with severe sepsis and septic shock: a meta-analysis of randomized controlled trials. *Journal of translational medicine* **16**, 1-14 (2018).

45      Zhang, L., Zhu, G., Han, L. & Fu, P. Early goal-directed therapy in the management of severe sepsis or septic shock in adults: a meta-analysis of randomized controlled trials. *BMC medicine* **13**, 1-12 (2015).

46      Kahn, G., Villaflor, A., Pong, V., Abbeel, P. & Levine, S. Uncertainty-aware reinforcement learning for collision avoidance. *arXiv preprint arXiv:1702.01182* (2017).

47      Osband, I., Blundell, C., Pritzel, A. & Van Roy, B. Deep exploration via bootstrapped DQN. *arXiv preprint arXiv:1602.04621* (2016).

48      Seymour, C. W. *et al.* Assessment of Clinical Criteria for Sepsis. *Jama* **315**, 762, doi:10.1001/jama.2016.0288 (2016).

49      Vincent, J. L. *et al.* The SOFA (Sepsis-related Organ Failure Assessment) score to describe organ dysfunction/failure. On behalf of the Working Group on Sepsis-Related Problems of the European Society of Intensive Care Medicine. *Intensive Care Med* **22**, 707-710 (1996).

50      Wilcoxon, F. Individual Comparisons by Ranking Methods. *Biometrics Bulletin* **1**, 80, doi:10.2307/3001968 (1945).